\documentclass[10pt,twocolumn,letterpaper]{article}

\usepackage[pagenumbers]{cvpr}   

\usepackage{booktabs}
\usepackage{tabularx}
\usepackage{array}
\usepackage{makecell}

\definecolor{cvprblue}{rgb}{0.21,0.49,0.74}
\usepackage[pagebackref,breaklinks,colorlinks,allcolors=cvprblue]{hyperref}

\def\paperID{*****}
\def\confName{CVPR}
\def\confYear{2026}

\newcolumntype{L}[1]{>{\raggedright\arraybackslash}p{#1}}
\newcolumntype{Y}{>{\raggedright\arraybackslash}X}

\title{CLIP-Guided Data Augmentation for Night-Time Image Dehazing}

\author{
Xining Ge$^{1}$ \quad
Weijun Yuan$^{2}$ \quad
Gengjia Chang$^{3}$ \quad
Xuyang Li$^{4}$ \quad
Shuhong Liu$^{5,\dag}$\\[0.5em]
$^{1}$Hangzhou Dianzi University \quad
$^{2}$Jinan University \quad
$^{3}$Hefei University of Technology\\
$^{4}$Wuhan University \quad
$^{5}$The University of Tokyo
}

\begin{document}
\maketitle

\begin{abstract}
Nighttime image dehazing faces a more complex degradation pattern than its daytime counterpart, as haze scattering couples with low illumination, non-uniform lighting, and strong light interference. Under limited supervision, this complexity aggravates domain drift and training instability, since target-domain samples are scarce while naively introducing external data may weaken adaptation due to distribution mismatch. This paper presents our solution to the NTIRE 2026 Night Time Image Dehazing Challenge, built as a unified framework that integrates domain-aligned data construction, stage-wise training, and inference-time enhancement. Specifically, a pre-trained CLIP visual encoder screens candidate external samples by similarity to construct training data closer to the target domain. NAFNet is then trained in two stages, first adapting to the target domain and then expanding to broader degradation patterns. At inference time, TLC, $\times$8 self-ensemble, and weighted snapshot fusion are combined to improve output stability. Rather than relying on complex network redesign, the proposed framework offers a practical and effective pipeline for nighttime image dehazing.
\end{abstract}


\section{Introduction}
Image restoration is a fundamental task in low-level vision that underpins a wide range of downstream applications, including autonomous driving~\cite{lisgs2025,liumg2025,zhou2024mod}, VR/AR \cite{lidense2025,ren2026esr}, 3D reconstruction~\cite{liu2025i2nerf,cui2026unifying,liu20263drr}, and scene understanding under adverse conditions~\cite{liuderain2025,liu2025realx3d,liudenoise2026}. As visual perception systems are increasingly deployed in complex real-world environments, the ability to recover high-quality images from degraded observations directly affects the reliability of subsequent recognition, navigation, and decision-making pipelines. Among many challenging scenarios, foggy nighttime environments stand out due to the combined effect of multiple degradation factors. Low illumination reduces the signal-to-noise ratio, non-uniform lighting and local strong light sources destabilize the brightness distribution, and haze scattering further attenuates contrast, suppresses detail, and distorts color. Unlike daytime dehazing, which primarily addresses scattering and transmittance estimation, nighttime degradation arises from the coupling of haze, low light, and artificial illumination, causing spatially varying failure modes such as halo diffusion near bright lights, noise accumulation in dark regions, and severe contrast loss at distant areas. The real challenge therefore lies in balancing structural recovery, detail preservation, and visual naturalness under highly heterogeneous lighting conditions.

Deep learning has substantially advanced image restoration. CNN-based architectures such as MIRNet, MPRNet, HINet, and NAFNet have demonstrated strong capability through multi-scale processing, feature interaction, and lightweight design~\cite{zamir2020learning,zamir2021multi,chen2021hinet,chen2022simple}. More recently, Transformer-based methods including SwinIR, Uformer, Restormer, and MAXIM further improve quality by strengthening long-range dependency modeling and global context reasoning~\cite{liang2021swinir,wang2022uformer,zamir2022restormer,tu2022maxim}. All-in-one restoration paradigms have also expanded model flexibility across diverse degradation types~\cite{chu2022improving,li2022all,guo2024onerestore}. Despite these advances, most methods assume relatively sufficient training data and stable degradation distributions. Nighttime image dehazing \cite{lin2025nighthaze} presents a fundamentally different regime where paired target-domain data are extremely limited, degradations are complex and heterogeneous, and the domain is highly unstable \cite{li2018benchmarking,ancuti2019dense}. Simply adopting a higher-capacity network does not necessarily help, as insufficient supervision may lead to unstable optimization. Introducing external data is a natural alternative, but without explicit control of domain discrepancy, distribution mismatch may weaken rather than improve adaptation. Recent challenge-driven studies further show that restoration under real-world adverse conditions is extending beyond conventional 2D enhancement to smoke-degraded and extreme low-light 3D scenes, where staged pipelines, enhancement priors, and physics-aware reconstruction have become increasingly common design choices~\cite{liu2026ntire,zheng20263d,liu2026elog,fu2026smokegs,cao2026gensmoke,zhu2026naka,guo2026reliability,chen2026dehaze}.

Based on these observations, this paper presents our solution to the NTIRE 2026 Night Time Image Dehazing Challenge \cite{ntire26nthaze_rep}. Rather than relying on complex network redesign, we construct a unified framework through task-oriented co-design of data screening, stage-wise training, and inference-time enhancement. At the data level, a pre-trained CLIP visual encoder evaluates candidate external samples by similarity to retain those closer to the target nighttime domain. At the optimization level, NAFNet is trained in two stages, first adapting to target-domain degradation and then extending to broader degradation patterns. At the inference level, TLC, self-ensemble, and snapshot ensemble are combined to improve output stability and reconstruction quality.

Our main contributions can be summarized as follows:
\begin{itemize}
    \item We adopt a domain-consistent data screening strategy based on pre-trained visual representations to mitigate target-domain data scarcity and domain shift from external data in nighttime image dehazing.
    \item We present a stage-wise restoration framework that jointly addresses data screening, training, and inference-time enhancement, demonstrating that task-oriented co-design of the training pipeline can be an effective route beyond simply increasing network complexity.
\end{itemize}

\section{Related Works}
\subsection{Single Image Dehazing}
Single image dehazing has long been an important research direction in image restoration. Most early methods were based on atmospheric scattering models and estimated key variables such as transmittance, scene irradiance, or air light by designing prior constraints \cite{narasimhan2002vision,narasimhan2000interactive}. Typical ideas include visibility priors \cite{tan2008visibility}, factorization-based single-image dehazing \cite{fattal2008single}, dark channel prior \cite{he2010single}, boundary constraint regularization \cite{meng2013efficient}, color attenuation prior \cite{zhu2015fast}, and non-local dehazing \cite{berman2016non}. This type of method has clear physical interpretability and achieves good results under moderate haze concentrations and relatively ideal imaging conditions. However, methods based on manual priors usually rely on strong scene assumptions, and their recovery effects degrade significantly when there are complex illumination changes, color imbalance, or non-uniform degradation.

With the development of deep learning, dehazing methods based on convolutional neural networks have gradually become mainstream. Related research directly establishes the mapping relationship between hazy and clear images through end-to-end learning, which significantly improves detail recovery capability and overall visual quality. Representative deep dehazing models include GridDehazeNet \cite{liu2019griddehazenet}, MSBDN \cite{dong2020multi}, FFA-Net \cite{qin2020ffa}, semi-supervised dehazing \cite{li2019semi}, transmission-aware dehazing transformers \cite{guo2022image}, vision-transformer-based dehazing \cite{song2023vision}, and recent CNN-attention hybrids such as DEA-Net \cite{chen2024dea} and depth-assisted dehazing \cite{zhang2024depth}. In recent years, with the widespread application of Transformer architectures in low-level vision, the global modeling capability of dehazing models has been further enhanced, leading to clear gains in complex scenes. However, the effectiveness of such methods usually relies on large-scale training data and relatively stable distributions, so their generalization ability still faces challenges when target-domain data are limited or the degradation distribution changes substantially.

\subsection{Nighttime Image Dehazing}
Compared with daytime dehazing, nighttime image dehazing faces a more complex imaging mechanism and stronger scene uncertainty. In nighttime environments, haze scattering couples with low illumination, non-uniform lighting, halo diffusion, noise amplification, and color shift, producing spatially varying degradation that is far more heterogeneous than daytime haze~\cite{li2015nighttime,liu2022nighttime,lin2025nighthaze}. Many daytime dehazing methods therefore transfer poorly to nighttime conditions, often failing at highlight suppression, dark-region detail recovery, and overall color naturalness. This problem also connects to the broader low-light enhancement literature, where decomposition-and-enhancement models~\cite{xu2020learning}, zero-reference enhancement~\cite{guo2020zero}, unpaired learning~\cite{jiang2021enlightengan}, Retinex-inspired designs~\cite{zhang2019kindling,wu2022uretinex,cai2023retinexformer}, zero-shot illumination-guided restoration~\cite{shi2024zero}, specularity-aware factorization~\cite{saini2024specularity}, and event-guided enhancement~\cite{liang2024towards} have been actively explored.

Existing nighttime dehazing methods improve performance mainly from two directions, namely introducing more detailed modeling of the nighttime imaging process to describe coupled degradation~\cite{li2015nighttime,liu2022nighttime}, and constructing more expressive deep networks that directly learn the degradation-to-clean mapping, with recent work further emphasizing non-homogeneous and data-efficient settings~\cite{liu2022towards,shetty2023non,shyam2023data,chen2024dea,zhang2024depth}. However, real nighttime degradation distributions are highly complex, target-domain samples are limited, and external data carry obvious domain gaps. The performance bottleneck therefore does not stem entirely from insufficient network expressiveness. Maintaining data-distribution consistency under limited supervision and improving training and inference stability are equally critical issues.

\subsection{Data Selection and Robust Restoration under Limited Supervision}
When target-domain samples are scarce, introducing external data is a common way to expand training data, but its effect depends largely on the distribution consistency between the external data and the target domain. Without effective filtering, additional data can increase the training scale but may also introduce significant domain shifts due to mismatches in scene content or degradation patterns, thereby weakening the model's learning performance on the target task. In recent years, pre-trained visual representations, including contrastive learning and vision-language pre-training \cite{chen2020simple,he2020momentum,caron2021emerging,radford2021learning,jia2021scaling,dosovitskiy2020image}, have shown strong capability for similar-sample selection and cross-domain transfer, providing new ideas for sample screening based on feature similarity. Compared with indiscriminately mixing external data, using general representations to select candidate samples helps expand the training distribution while maintaining target-domain relevance.

Related evidence from adjacent restoration tasks also suggests that performance gains do not always come from heavier backbone redesign. Stronger data-centric training can substantially improve restoration quality even without introducing a new heavy architecture~\cite{chang2026beyond}. In adjacent super-resolution settings, TLC-style local conversion, self-ensemble, and training-free output fusion have also been used to improve robustness at inference time~\cite{ge2026dual,chang2026training}. These observations reinforce the broader view that, under limited supervision and unstable degradations, careful coordination of data organization, optimization, and inference can be as important as the backbone itself.

On the other hand, in small-sample or high-uncertainty scenarios, restoration performance depends not only on the network learning process during training, but also on the stability of the results at inference time. Strategies such as test-time training \cite{liu2022towards} and model ensemble \cite{huang2017snapshot} have been widely used to improve prediction robustness in low-level vision tasks. These methods usually do not change the basic network structure, but integrate prediction results from multiple views to alleviate fluctuations caused by a single inference and thus obtain more stable outputs. For tasks such as nighttime image dehazing, which involve complex degradation and limited target-domain supervision, effectively combining domain-consistency-oriented data organization, robust training, and inference enhancement remains a promising direction.


\begin{figure*}[!tp]
    \centering
    \includegraphics[width=0.94\textwidth]{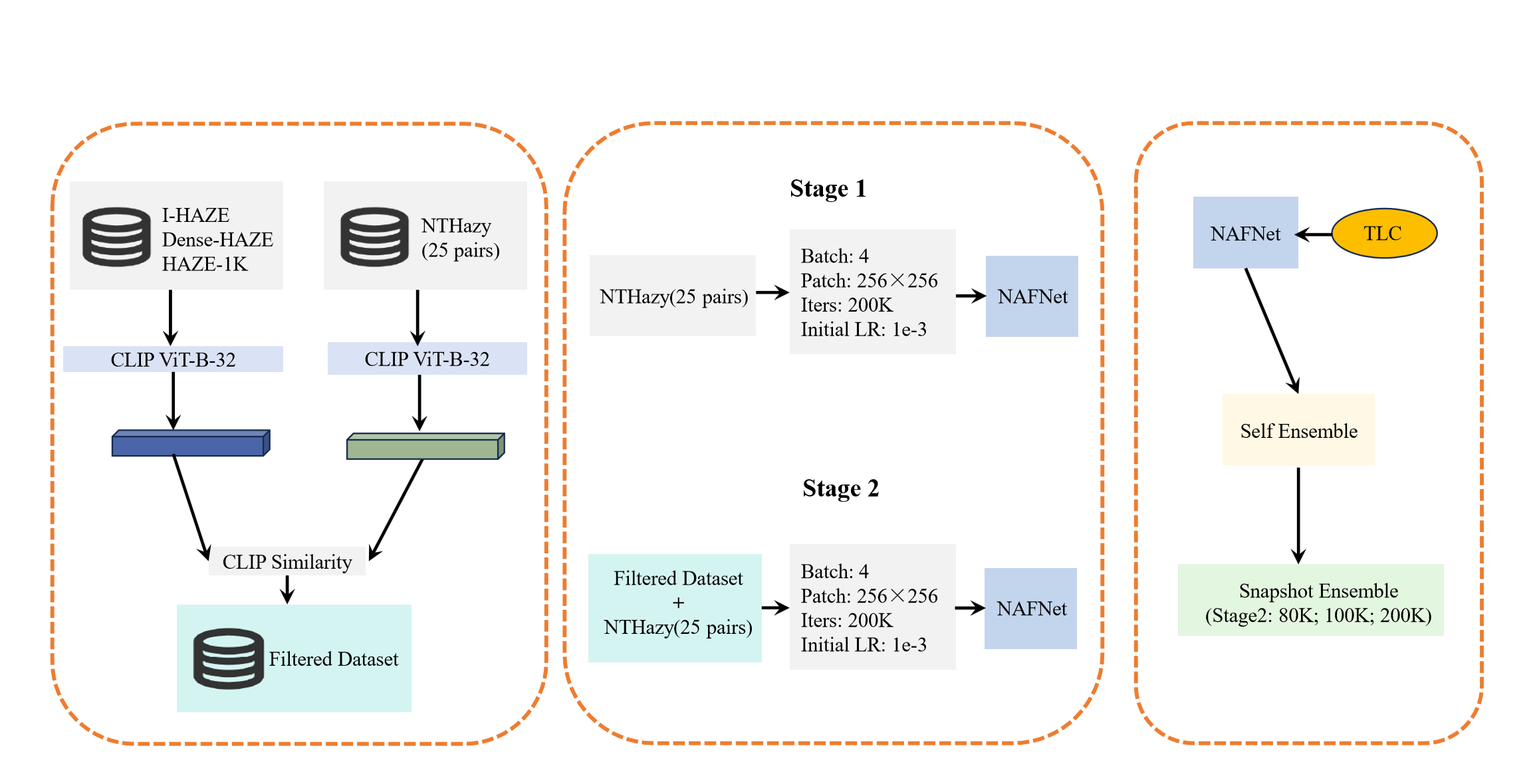}
    \caption{Pipeline overall pipeline of our proposed solution, showing CLIP-guided data augmentation, stage-wise NAFNet training, and inference-time enhancement.}
    \label{fig:pipeline}
\end{figure*}

\section{Method}
\subsection{Framework Overview}
Our framework comprises three tightly coupled components: target-aligned auxiliary data curation, stage-wise restoration training, and inference-time enhancement. As illustrated in Figure~\ref{fig:pipeline}, the pipeline proceeds from auxiliary data screening, through two-stage training, to test-time enhancement. Candidate external samples are first compared with target nighttime images through a pre-trained CLIP visual encoder~\cite{radford2021learning}, and only those with sufficiently high semantic similarity are retained to form an augmented training set. NAFNet~\cite{chen2022simple} is then optimized in two stages, so that it first captures target-domain characteristics and subsequently absorbs the filtered auxiliary data in a more stable manner. At inference, TLC-style global information aggregation~\cite{chu2022improving}, $\times 8$ self-ensemble, and weighted snapshot fusion are combined to improve output robustness and reconstruction quality. Rather than relying on heavy backbone redesign, the framework improves nighttime dehazing through a coordinated design of data organization, optimization strategy, and test-time integration.

\subsection{Cross-Dataset Data Curation}
The NTHazy target domain provides only a limited number of paired nighttime samples, which is insufficient for training a high-capacity restoration model. A natural remedy is to draw on existing real dehazing datasets such as I-HAZE, Dense-Haze, and HAZE1K. These benchmarks, however, are dominated by daytime or otherwise non-nighttime captures, and indiscriminately mixing them with NTHazy would widen the gap between the training distribution and the nighttime test distribution, ultimately weakening the model's ability to characterize night-specific degradations. We therefore frame auxiliary data usage not as unconstrained sample expansion, but as the construction of an extended training set whose distribution remains aligned with the nighttime target domain.
We treat NTHazy as the in-domain paired training set and regard I-HAZE, Dense-Haze, and HAZE1K as candidate auxiliary sources. Real paired hazy/clear benchmarks of this kind, together with the NTIRE dehazing series, have repeatedly been shown to be valuable for robust evaluation and model development~\cite{ancuti2018haze,ancuti2018haze,ancuti2020nh,ancuti2020ntire,ancuti2021ntire,ancuti2024ntire}, as they supply genuine degradation patterns and rich textural diversity. Their imaging conditions nevertheless differ substantially from nighttime scenes, so each external sample is screened for its relevance to the target domain before being admitted into training.
For this screening we use the pre-trained CLIP ViT-B/32 visual encoder~\cite{radford2021learning}. Feature representations are extracted for both the NTHazy nighttime images and every candidate external image, and a semantic similarity is computed between each candidate and the nighttime target domain. Samples retained under this criterion are those that lie closer to nighttime captures in visual content, illumination, or degradation morphology, which limits cross-domain disturbance during training.
Under this criterion, the retained external data consist of 21 pairs from I-HAZE, 10 pairs from Dense-Haze, and 3 pairs from HAZE1K. Together with the 25 original NTHazy pairs, they form an extended training set of 59 pairs. Compared with naively mixing all external samples, this selective strategy preserves a meaningful degree of diversity while keeping the training distribution consistent with the nighttime target domain.

\subsection{Stage-Wise Restoration Training}
Even after similarity filtering, residual distributional gaps between the external samples and real nighttime scenes are unavoidable. Mixing all data from the very beginning would bias the model toward the larger but imperfectly aligned external set, weakening its modeling of the target nighttime degradation. We therefore adopt a stage-wise training scheme that first establishes a target-domain prior and then incorporates the filtered auxiliary data on top of it.

In the first stage, only the 25 nighttime pairs from NTHazy are used to train the restoration backbone. This allows the model to acquire a basic representation of nighttime haze degradation, low-light imaging characteristics, and target-domain texture statistics. Training is performed with a patch size of 256$\times$256, a batch size of 4, and $200$K iterations.

In the second stage, the CLIP-filtered extended set is introduced and the stage-one model is fine-tuned on the combined data for another $200$K iterations under the same patch size and batch size. Because the network has already internalized a preliminary nighttime prior, exposing it to semantically closer external samples broadens its coverage of degradation modes without overriding the target-domain behavior learned in the first stage. The two stages thus form a coherent optimization process in which target-domain pre-adaptation precedes, and constrains, the absorption of filtered auxiliary data.

\begin{figure*}[t]
    \centering
    \includegraphics[width=\textwidth]{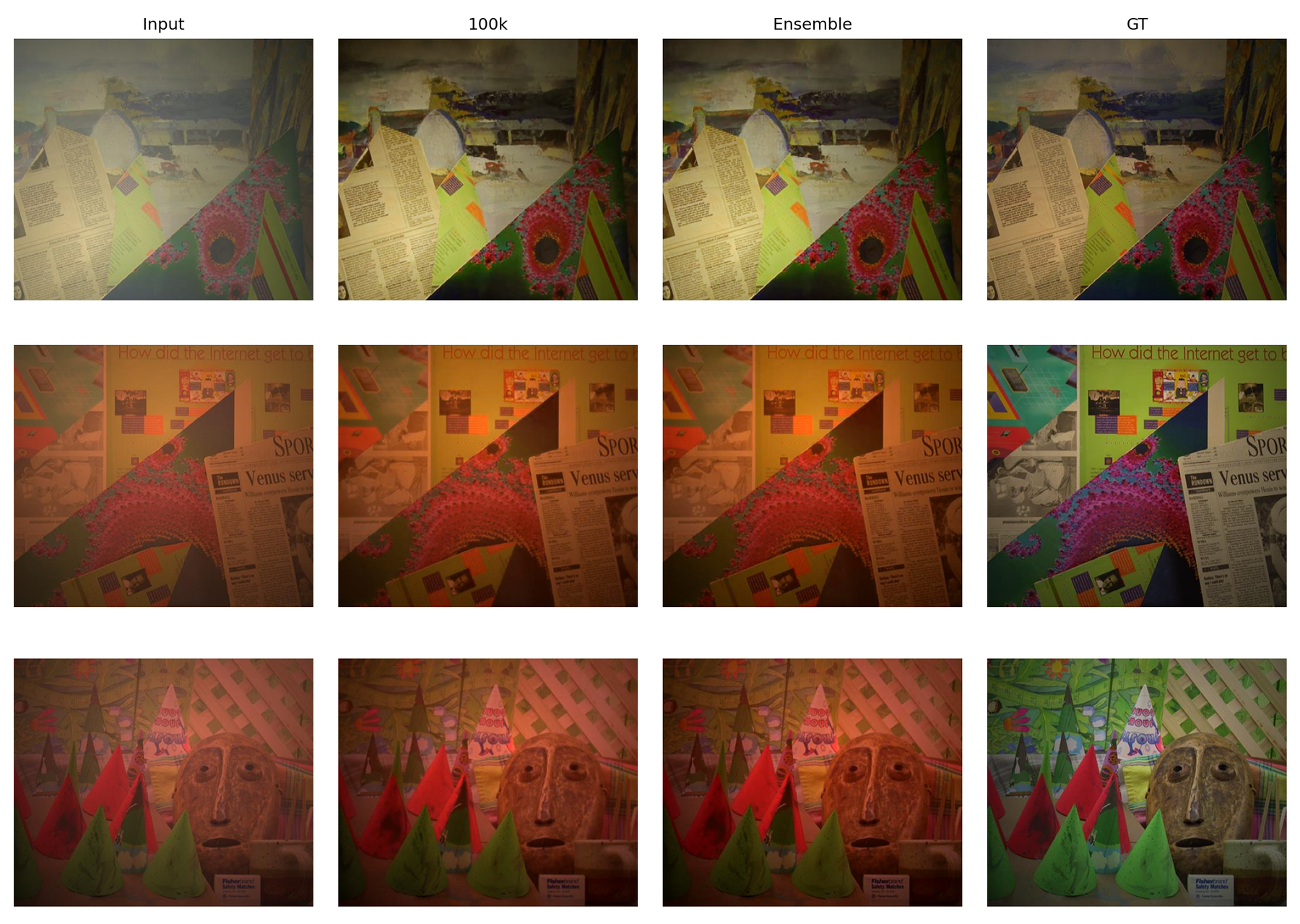}
    \caption{Qualitative comparison on representative NHM-20 samples. From left to right: Input, NAFNet-100k, Weighted Ensemble, and ground truth.}
    \label{fig:qualitative}
\end{figure*}

\subsection{Inference-Time Enhancement}

Beyond training-time data organization, we apply three complementary strategies at inference to improve robustness without modifying learned parameters.

We adopt NAFNetLocal~\cite{chen2022simple} as the inference network with TLC-style local enhancement~\cite{chu2022improving} enabled at initialization, which reconciles local statistics with global context and stabilizes restoration under complex illumination and fine textures. Inference is performed in a full-image forward pass by default, with an overlap-based tiled variant available for memory-constrained settings.

For $\times 8$ self-ensemble, the input is transformed under the eight combinations of vertical flip, horizontal flip, and transpose; predictions are mapped back to the original coordinates and averaged, improving stability at no training cost.

Finally, instead of relying on a single checkpoint, we fuse three snapshots from the training stage by ensembling~\cite{huang2017snapshot}. The restorations produced by the $80$K, $100$K, and $200$K checkpoints are linearly combined in the image domain with weights $0.04$, $0.01$, and $0.95$. The dominant weight on the final checkpoint reflects its best fit to the target domain, while the earlier snapshots act as mild regularizers against checkpoint-specific artifacts.

\section{Experiments}

\paragraph{Implementation Details.}
The restoration backbone is NAFNetLocal/NAFNet~\cite{chen2022simple} with width 32, encoder blocks $[1,1,1,28]$, one middle block, and decoder blocks $[1,1,1,1]$. Training follows a two-stage setup: the first stage uses only NTHazy for target-domain adaptation, and the second stage introduces CLIP-filtered auxiliary data for continued training. Both stages use batch size 4, patch size $256\times256$, 200K iterations, AdamW optimizer with initial learning rate $1\times10^{-3}$ and cosine annealing, MSE loss, and augmentation of random horizontal flipping and rotation. At inference time, inputs are cropped to multiples of 8, and three enhancements are applied: TLC-style local conversion, $\times$8 self-ensemble, and weighted snapshot fusion over the 80K, 100K, and 200K checkpoints. EMA weights (\texttt{params\_ema}) are loaded when available.

\begin{table}[t]
    \centering
    \small
    \setlength{\tabcolsep}{4pt}
    \renewcommand{\arraystretch}{1.12}
    \caption{Composition of the training data.}
    \label{tab:data-composition}
    \begin{tabularx}{\columnwidth}{L{0.28\columnwidth} L{0.16\columnwidth} Y}
        \toprule
        Data Source & \# Samples & Description \\
        \midrule
        NTHazy & 25 & Target-domain paired data \\
        I-HAZE (selected) & 21 & External supplementary data \\
        Dense-Haze (selected) & 10 & External supplementary data \\
        HAZE1K (selected) & 3 & External supplementary data \\
        Total & 59 & Selected training set size \\
        \bottomrule
    \end{tabularx}
\end{table}

\begin{table*}[t]
    \centering
    \small
    \setlength{\tabcolsep}{5pt}
    \renewcommand{\arraystretch}{1.12}
    \caption{Summary of the implementation and inference settings.}
    \label{tab:settings}
    \begin{tabularx}{\textwidth}{L{0.22\textwidth} Y}
        \toprule
        Item & Setting \\
        \midrule
        Basic network & NAFNetLocal + NAFNet \\
        Public inference structure & Width = 32, encoder blocks $[1, 1, 1, 28]$, one middle block, decoder blocks $[1, 1, 1, 1]$ \\
        Input preprocessing & Crop to multiples of 8 before inference; NHM public reproducible evaluation inputs are uniformly saved as PNG \\
        Stage I data & NTHazy (25 pairs) \\
        Stage II data & NTHazy + filtered I-HAZE / Dense-Haze / HAZE1K (59 pairs in total) \\
        Training configuration & Batch size = 4, patch size = $256\times256$, 200K iterations per stage \\
        Optimization settings & AdamW, initial learning rate $1\times10^{-3}$, cosine annealing, MSE loss \\
        TLC / local conversion & Accessed via \texttt{Local\_Base.convert(...)} in NAFNetLocal \\
        Self-ensemble & x8; vertical / horizontal / transpose and combined transforms, followed by mean fusion \\
        Snapshot fusion & 80K / 100K / 200K three checkpoints, weighted fusion in image space \\
        Fusion weight & 0.04 / 0.01 / 0.95 \\
        \bottomrule
    \end{tabularx}
\end{table*}

\begin{table*}[t]
    \centering
    \small
    \setlength{\tabcolsep}{7pt}
    \renewcommand{\arraystretch}{1.10}
    \caption{Results on the NHM-20 public reproducible evaluation.}
    \label{tab:quantitative}
    \begin{tabular}{lccccc}
        \toprule
        Method & PSNR-Y $\uparrow$ & SSIM-Y $\uparrow$ & PSNR-RGB $\uparrow$ & SSIM-RGB $\uparrow$ & LPIPS-Alex $\downarrow$ \\
        \midrule
        Input & 23.4697 & 0.887581 & 19.4925 & 0.811362 & 0.278838 \\
        NAFNet-80k & 24.6568 & 0.906439 & 18.2269 & 0.776861 & 0.310485 \\
        NAFNet-100k & 25.0125 & 0.904238 & 18.7441 & 0.782752 & 0.311571 \\
        NAFNet-200k & 24.8830 & 0.904796 & 18.3753 & 0.763607 & 0.318563 \\
        Weighted Ensemble & 24.8875 & 0.905079 & 18.3813 & 0.764708 & 0.317944 \\
        \bottomrule
    \end{tabular}
\end{table*}

\paragraph{Datasets.}
The training set comprises NTHazy target-domain pairs and CLIP-filtered auxiliary samples from I-HAZE, Dense-Haze, and HAZE1K, totaling 59 paired images. Table~\ref{tab:data-composition} summarizes the composition. For public reproducible evaluation, we construct NHM-20 from 20 aligned image pairs (four per haze level from 1 to 5), with inputs from \texttt{img\_0} and ground truth from \texttt{img\_1}, both cropped to multiples of 8 and saved as PNG to avoid JPG compression instability.

\paragraph{Evaluation Protocol.}
The official NTIRE 2026 hidden-test result is reported as the competition outcome. The NHM-20 public evaluation provides a controlled, reproducible setting for analyzing data composition and metric behavior. Following the NHM convention, Y-channel PSNR/SSIM serve as the primary metrics, while RGB-channel metrics and LPIPS are reported as extra references. 

\subsection{Quantitative and Qualitative Analysis}
Table~\ref{tab:quantitative} reports results on NHM-20. The public team07 pipeline improves Y-channel PSNR and SSIM over the hazy-input baseline, confirming effectiveness in recovering brightness structure. However, RGB metrics and LPIPS do not surpass the input baseline, suggesting that the current pipeline primarily benefits luminance-domain restoration while color fidelity and perceptual quality require further attention. Regarding ensemble behavior, the best PSNR-Y is achieved by the 100K checkpoint and the best SSIM-Y by the 80K checkpoint; the weighted ensemble stabilizes output quality across diverse scenes but does not guarantee per-metric optimality over every single checkpoint, indicating that adaptive checkpoint selection may offer further improvement. It is also worth noting that the CLIP-guided data filtering and two-stage training process are not directly verifiable from the public code; the corresponding training-side description is based on the solution factsheet.

Figure~\ref{fig:qualitative} shows visual comparisons on NHM-20. Overall, the combination of domain-consistent data filtering, stage-wise training, and inference-time enhancement constitutes a complete and reproducible technical route for small-sample nighttime dehazing, with stable brightness-structure recovery confirmed by the NHM-20 evaluation. Perceptual-level improvement remains scene-dependent and warrants further investigation.

\section{Conclusion}
This paper presents a unified framework for nighttime image dehazing that integrates domain-consistent data screening, stage-wise training, and inference-time enhancement to address target-domain data scarcity and prediction instability. At the training level, CLIP-guided sample filtering and two-stage optimization improve target-domain adaptability. At the inference level, TLC-style local conversion, $\times$8 self-ensemble, and weighted snapshot fusion form a robust inference pipeline. Beyond the official NTIRE hidden-test result, a public reproducible evaluation on NHM-20 confirms that the method achieves stable Y-channel PSNR/SSIM improvement over the hazy-input baseline, while RGB metrics and LPIPS do not exceed the baseline, indicating that the main benefit lies in brightness-structure recovery rather than uniform perceptual improvement. Future work may focus on more refined cross-domain sample screening and degradation modeling closer to real nighttime imaging.

{\small
\bibliographystyle{plain}
\bibliography{references}

@article{narasimhan2002vision,
  title={Vision and the atmosphere},
  author={Narasimhan, Srinivasa G and Nayar, Shree K},
  journal={International journal of computer vision},
  volume={48},
  number={3},
  pages={233--254},
  year={2002},
  publisher={Springer}
}

@article{narasimhan2000interactive,
  title={Interactive (de) weathering of an image using physical models},
  author={Narasimhan, Srinivasa},
  year={2000}
}

@inproceedings{lin2025nighthaze,
  title={Nighthaze: Nighttime image dehazing via self-prior learning},
  author={Lin, Beibei and Jin, Yeying and Wending, Yan and Ye, Wei and Yuan, Yuan and Tan, Robby T},
  booktitle={Proceedings of the AAAI Conference on Artificial Intelligence},
  volume={39},
  number={5},
  pages={5209--5217},
  year={2025}
}

@article{liuderain2025,
    author={Liu, Shuhong and Chen, Xiang and Chen, Hongming and Xu, Quanfeng and Li, Mingrui},
    title={DeRainGS: Gaussian Splatting for Enhanced Scene Reconstruction in Rainy Environments},
    volume={39},
    DOI={10.1609/aaai.v39i5.32592},
    number={5},
    journal={Proceedings of the AAAI Conference on Artificial Intelligence},
    year={2025},
    pages={5558-5566}
}

@inproceedings{liu2025i2nerf,
    title={I2-NeRF: Learning Neural Radiance Fields Under Physically-Grounded Media Interactions},
    author={Liu, Shuhong and Gu, Lin and Cui, Ziteng and Chu, Xuangeng and Harada, Tatsuya},
    booktitle={Advances in Neural Information Processing Systems},
    year={2025},
}

@inproceedings{lisgs2025,
    title={SGS-SLAM: Semantic Gaussian Splatting for Neural Dense SLAM},
    author={Li, Mingrui and Liu, Shuhong and Zhou, Heng and Zhu, Guohao and Cheng, Na and Deng, Tianchen and Wang, Hongyu},
    booktitle={European Conference on Computer Vision},
    year={2025},
    pages={163--179},
}

@article{liumg2025,
    title={MG-SLAM: Structure Gaussian Splatting SLAM With Manhattan World Hypothesis}, 
    author={Liu, Shuhong and Deng, Tianchen and Zhou, Heng and Li, Liuzhuozheng and Wang, Hongyu and Wang, Danwei and Li, Mingrui},
    journal={IEEE Transactions on Automation Science and Engineering}, 
    year={2025},
    volume={22},
    number={},
    pages={17034-17049},
    doi={10.1109/TASE.2025.3575772}
}

@inproceedings{ntire26nthaze_rep, 
    title={{NTIRE 2026 Nighttime Image Dehazing Challenge Report}}, 
    author={Ancuti, Radu and  Brateanu, Alexandru and  Vasluianu, Florin and  Balmez, Raul and  Orhei, Ciprian and  Ancuti, Codruta and  Timofte, Radu and  Ancuti, Cosmin and others},   
    booktitle={Proceedings of the IEEE/CVF Conference on Computer Vision and Pattern Recognition (CVPR) Workshops},  
    year = {2026} 
}

@article{lidense2025,
    title={DenseSplat: Densifying Gaussian Splatting SLAM with Neural Radiance Prior},
    author={Li, Mingrui and Liu, Shuhong and Deng, Tianchen and Wang, Hongyu},
    journal={IEEE Transactions on Visualization \& Computer Graphics},
    year={2025},
    volume={},
    number={01},
    ISSN={1941-0506},
    pages={1-14},
    doi={10.1109/TVCG.2025.3617961},
    publisher={IEEE Computer Society},
}

@article{liu2025realx3d,
    title={RealX3D: A Physically-Degraded 3D Benchmark for Multi-view Visual Restoration and Reconstruction},
    author={Liu, Shuhong and Bao, Chenyu and Cui, Ziteng and Liu, Yun and Chu, Xuangeng and Gu, Lin and Conde, Marcos V and Umagami, Ryo and Hashimoto, Tomohiro and Hu, Zijian and others},
    journal={arXiv preprint arXiv:2512.23437},
    year={2026}
}

@article{liudenoise2026,
    title={Denoising the Deep Sky: Physics-Based CCD Noise Formation for Astronomical Imaging},
    author={Liu, Shuhong and Ge, Xining and Gu, Ziying and Gu, Lin and Cui, Ziteng and Chu, Xuangeng and Liu, Jun and Li, Dong and Harada, Tatsuya},
    journal={arXiv preprint arXiv:2601.23276},
    year={2026}
}

@article{li2018benchmarking,
  title={Benchmarking single-image dehazing and beyond},
  author={Li, Boyi and Ren, Wenqi and Fu, Dengpan and Tao, Dacheng and Feng, Dan and Zeng, Wenjun and Wang, Zhangyang},
  journal={IEEE transactions on image processing},
  volume={28},
  number={1},
  pages={492--505},
  year={2018},
  publisher={IEEE}
}

@inproceedings{ancuti2019dense,
  title={Dense-haze: A benchmark for image dehazing with dense-haze and haze-free images},
  author={Ancuti, Codruta O and Ancuti, Cosmin and Sbert, Mateu and Timofte, Radu},
  booktitle={2019 IEEE international conference on image processing (ICIP)},
  pages={1014--1018},
  year={2019},
  organization={IEEE}
}

@inproceedings{ancuti2018haze,
  title={O-haze: a dehazing benchmark with real hazy and haze-free outdoor images},
  author={Ancuti, Codruta O and Ancuti, Cosmin and Timofte, Radu and De Vleeschouwer, Christophe},
  booktitle={Proceedings of the IEEE conference on computer vision and pattern recognition workshops},
  pages={754--762},
  year={2018}
}

@article{cui2026unifying,
  title={Unifying Color and Lightness Correction with View-Adaptive Curve Adjustment for Robust 3D Novel View Synthesis},
  author={Cui, Ziteng and Liu, Shuhong and Dong, Xiaoyu and Chu, Xuangeng and Gu, Lin and Yang, Ming-Hsuan and Harada, Tatsuya},
  journal={arXiv preprint arXiv:2602.18322},
  year={2026}
}

@article{zhou2024mod,
  title={Mod-slam: Monocular dense mapping for unbounded 3d scene reconstruction},
  author={Zhou, Heng and Guo, Zhetao and Ren, Yuxiang and Liu, Shuhong and Zhang, Lechen and Zhang, Kaidi and Li, Mingrui},
  journal={IEEE Robotics and Automation Letters},
  volume={10},
  number={1},
  pages={484--491},
  year={2024},
  publisher={IEEE}
}

@inproceedings{tan2008visibility,
  title={Visibility in bad weather from a single image},
  author={Tan, Robby T},
  booktitle={2008 IEEE conference on computer vision and pattern recognition},
  pages={1--8},
  year={2008},
  organization={IEEE}
}

@article{fattal2008single,
  title={Single image dehazing},
  author={Fattal, Raanan},
  journal={ACM transactions on graphics (TOG)},
  volume={27},
  number={3},
  pages={1--9},
  year={2008},
  publisher={ACM New York, NY, USA}
}

@article{he2010single,
  title={Single image haze removal using dark channel prior},
  author={He, Kaiming and Sun, Jian and Tang, Xiaoou},
  journal={IEEE transactions on pattern analysis and machine intelligence},
  volume={33},
  number={12},
  pages={2341--2353},
  year={2010},
  publisher={Ieee}
}

@inproceedings{meng2013efficient,
  title={Efficient image dehazing with boundary constraint and contextual regularization},
  author={Meng, Gaofeng and Wang, Ying and Duan, Jiangyong and Xiang, Shiming and Pan, Chunhong},
  booktitle={Proceedings of the IEEE international conference on computer vision},
  pages={617--624},
  year={2013}
}

@article{zhu2015fast,
  title={A fast single image haze removal algorithm using color attenuation prior},
  author={Zhu, Qingsong and Mai, Jiaming and Shao, Ling},
  journal={IEEE transactions on image processing},
  volume={24},
  number={11},
  pages={3522--3533},
  year={2015},
  publisher={IEEE}
}

@inproceedings{berman2016non,
  title={Non-local image dehazing},
  author={Berman, Dana and Avidan, Shai and others},
  booktitle={Proceedings of the IEEE conference on computer vision and pattern recognition},
  pages={1674--1682},
  year={2016}
}

@inproceedings{ancuti2020nh,
  title={NH-HAZE: An image dehazing benchmark with non-homogeneous hazy and haze-free images},
  author={Ancuti, Codruta O and Ancuti, Cosmin and Timofte, Radu},
  booktitle={Proceedings of the IEEE/CVF conference on computer vision and pattern recognition workshops},
  pages={444--445},
  year={2020}
}

@inproceedings{ancuti2020ntire,
  title={Ntire 2020 challenge on nonhomogeneous dehazing},
  author={Ancuti, Codruta O and Ancuti, Cosmin and Vasluianu, Florin-Alexandru and Timofte, Radu},
  booktitle={Proceedings of the IEEE/CVF conference on computer vision and pattern recognition workshops},
  pages={490--491},
  year={2020}
}

@inproceedings{ancuti2021ntire,
  title={NTIRE 2021 nonhomogeneous dehazing challenge report},
  author={Ancuti, Codruta O and Ancuti, Cosmin and Vasluianu, Florin-Alexandru and Timofte, Radu},
  booktitle={Proceedings of the IEEE/CVF Conference on Computer Vision and Pattern Recognition},
  pages={627--646},
  year={2021}
}

@inproceedings{ancuti2024ntire,
  title={NTIRE 2024 dense and non-homogeneous dehazing challenge report},
  author={Ancuti, Codruta O and Ancuti, Cosmin and Vasluianu, Florin-Alexandru and Timofte, Radu and Liu, Yidi and Wang, Xingbo and Zhu, Yurui and Shi, Gege and Lu, Xin and Fu, Xueyang and others},
  booktitle={Proceedings of the IEEE/CVF Conference on Computer Vision and Pattern Recognition},
  pages={6453--6468},
  year={2024}
}

@inproceedings{dong2020multi,
  title={Multi-scale boosted dehazing network with dense feature fusion},
  author={Dong, Hang and Pan, Jinshan and Xiang, Lei and Hu, Zhe and Zhang, Xinyi and Wang, Fei and Yang, Ming-Hsuan},
  booktitle={Proceedings of the IEEE/CVF conference on computer vision and pattern recognition},
  pages={2157--2167},
  year={2020}
}

@inproceedings{qin2020ffa,
  title={FFA-Net: Feature fusion attention network for single image dehazing},
  author={Qin, Xu and Wang, Zhilin and Bai, Yuanchao and Xie, Xiaodong and Jia, Huizhu},
  booktitle={Proceedings of the AAAI conference on artificial intelligence},
  volume={34},
  number={07},
  pages={11908--11915},
  year={2020}
}

@article{li2019semi,
  title={Semi-supervised image dehazing},
  author={Li, Lerenhan and Dong, Yunlong and Ren, Wenqi and Pan, Jinshan and Gao, Changxin and Sang, Nong and Yang, Ming-Hsuan},
  journal={IEEE Transactions on Image Processing},
  volume={29},
  pages={2766--2779},
  year={2019},
  publisher={IEEE}
}

@inproceedings{liu2022towards,
  title={Towards multi-domain single image dehazing via test-time training},
  author={Liu, Huan and Wu, Zijun and Li, Liangyan and Salehkalaibar, Sadaf and Chen, Jun and Wang, Keyan},
  booktitle={Proceedings of the IEEE/CVF conference on computer vision and pattern recognition},
  pages={5831--5840},
  year={2022}
}

@inproceedings{guo2022image,
  title={Image dehazing transformer with transmission-aware 3d position embedding},
  author={Guo, Chun-Le and Yan, Qixin and Anwar, Saeed and Cong, Runmin and Ren, Wenqi and Li, Chongyi},
  booktitle={Proceedings of the IEEE/CVF conference on computer vision and pattern recognition},
  pages={5812--5820},
  year={2022}
}

@article{song2023vision,
  title={Vision transformers for single image dehazing},
  author={Song, Yuda and He, Zhuqing and Qian, Hui and Du, Xin},
  journal={IEEE Transactions on Image Processing},
  volume={32},
  pages={1927--1941},
  year={2023},
  publisher={IEEE}
}

@inproceedings{liu2022nighttime,
  title={Nighttime image dehazing based on variational decomposition model},
  author={Liu, Yun and Yan, Zhongsheng and Wu, Aimin and Ye, Tian and Li, Yuche},
  booktitle={Proceedings of the IEEE/CVF conference on computer vision and pattern recognition},
  pages={640--649},
  year={2022}
}

@inproceedings{shetty2023non,
  title={Non homogeneous realistic single image dehazing},
  author={Shetty, Lithesh and others},
  booktitle={Proceedings of the IEEE/CVF Winter Conference on Applications of Computer Vision},
  pages={548--555},
  year={2023}
}

@inproceedings{shyam2023data,
  title={Data efficient single image dehazing via adversarial auto-augmentation and extended atmospheric scattering model},
  author={Shyam, Pranjay and Yoo, HyunJin},
  booktitle={Proceedings of the IEEE/CVF international conference on computer vision},
  pages={227--237},
  year={2023}
}

@article{chen2024dea,
  title={DEA-Net: Single image dehazing based on detail-enhanced convolution and content-guided attention},
  author={Chen, Zixuan and He, Zewei and Lu, Zhe-Ming},
  journal={IEEE transactions on image processing},
  volume={33},
  pages={1002--1015},
  year={2024},
  publisher={IEEE}
}

@inproceedings{zhang2024depth,
  title={Depth information assisted collaborative mutual promotion network for single image dehazing},
  author={Zhang, Yafei and Zhou, Shen and Li, Huafeng},
  booktitle={Proceedings of the IEEE/CVF conference on computer vision and pattern recognition},
  pages={2846--2855},
  year={2024}
}

@inproceedings{liu2019griddehazenet,
  title={Griddehazenet: Attention-based multi-scale network for image dehazing},
  author={Liu, Xiaohong and Ma, Yongrui and Shi, Zhihao and Chen, Jun},
  booktitle={Proceedings of the IEEE/CVF international conference on computer vision},
  pages={7314--7323},
  year={2019}
}

@inproceedings{li2015nighttime,
  title={Nighttime haze removal with glow and multiple light colors},
  author={Li, Yu and Tan, Robby T and Brown, Michael S},
  booktitle={Proceedings of the IEEE international conference on computer vision},
  pages={226--234},
  year={2015}
}

@inproceedings{xu2020learning,
  title={Learning to restore low-light images via decomposition-and-enhancement},
  author={Xu, Ke and Yang, Xin and Yin, Baocai and Lau, Rynson WH},
  booktitle={Proceedings of the IEEE/CVF conference on computer vision and pattern recognition},
  pages={2281--2290},
  year={2020}
}

@inproceedings{guo2020zero,
  title={Zero-reference deep curve estimation for low-light image enhancement},
  author={Guo, Chunle and Li, Chongyi and Guo, Jichang and Loy, Chen Change and Hou, Junhui and Kwong, Sam and Cong, Runmin},
  booktitle={Proceedings of the IEEE/CVF conference on computer vision and pattern recognition},
  pages={1780--1789},
  year={2020}
}

@article{jiang2021enlightengan,
  title={Enlightengan: Deep light enhancement without paired supervision},
  author={Jiang, Yifan and Gong, Xinyu and Liu, Ding and Cheng, Yu and Fang, Chen and Shen, Xiaohui and Yang, Jianchao and Zhou, Pan and Wang, Zhangyang},
  journal={IEEE transactions on image processing},
  volume={30},
  pages={2340--2349},
  year={2021},
  publisher={IEEE}
}

@inproceedings{zhang2019kindling,
  title={Kindling the darkness: A practical low-light image enhancer},
  author={Zhang, Yonghua and Zhang, Jiawan and Guo, Xiaojie},
  booktitle={Proceedings of the 27th ACM international conference on multimedia},
  pages={1632--1640},
  year={2019}
}

@inproceedings{wu2022uretinex,
  title={Uretinex-net: Retinex-based deep unfolding network for low-light image enhancement},
  author={Wu, Wenhui and Weng, Jian and Zhang, Pingping and Wang, Xu and Yang, Wenhan and Jiang, Jianmin},
  booktitle={Proceedings of the IEEE/CVF conference on computer vision and pattern recognition},
  pages={5901--5910},
  year={2022}
}

@inproceedings{cai2023retinexformer,
  title={Retinexformer: One-stage retinex-based transformer for low-light image enhancement},
  author={Cai, Yuanhao and Bian, Hao and Lin, Jing and Wang, Haoqian and Timofte, Radu and Zhang, Yulun},
  booktitle={Proceedings of the IEEE/CVF international conference on computer vision},
  pages={12504--12513},
  year={2023}
}

@inproceedings{shi2024zero,
  title={ZERO-IG: Zero-shot illumination-guided joint denoising and adaptive enhancement for low-light images},
  author={Shi, Yiqi and Liu, Duo and Zhang, Liguo and Tian, Ye and Xia, Xuezhi and Fu, Xiaojing},
  booktitle={Proceedings of the IEEE/CVF conference on computer vision and pattern recognition},
  pages={3015--3024},
  year={2024}
}

@inproceedings{saini2024specularity,
  title={Specularity factorization for low-light enhancement},
  author={Saini, Saurabh and Narayanan, PJ},
  booktitle={Proceedings of the IEEE/CVF Conference on Computer Vision and Pattern Recognition},
  pages={1--12},
  year={2024}
}

@inproceedings{liang2024towards,
  title={Towards robust event-guided low-light image enhancement: a large-scale real-world event-image dataset and novel approach},
  author={Liang, Guoqiang and Chen, Kanghao and Li, Hangyu and Lu, Yunfan and Wang, Lin},
  booktitle={Proceedings of the IEEE/CVF Conference on Computer Vision and Pattern Recognition},
  pages={23--33},
  year={2024}
}

@article{ren2026esr,
  title={The Eleventh {NTIRE} 2026 Efficient Super-Resolution Challenge Report},
  author={Ren, Bin and Guo, Hang and Shu, Yan and Ma, Jiaqi and Cui, Ziteng and Liu, Shuhong and Mei, Guofeng and Sun, Lei and Wu, Zongwei and Khan, Fahad Shahbaz and Khan, Salman and Timofte, Radu and Li, Yawei and others},
  journal={arXiv preprint arXiv:2604.03198},
  year={2026}
}

@article{liu20263drr,
  title={NTIRE 2026 {3D} Restoration and Reconstruction in Adverse Conditions: {RealX3D} Challenge Results},
  author={Liu, Shuhong and Bao, Chenyu and Cui, Ziteng and Chu, Xuangeng and Ren, Bin and Gu, Lin and Chen, Xiang and Li, Mingrui and Ma, Long and Conde, Marcos V. and Timofte, Radu and others},
  journal={arXiv preprint arXiv:2604.04135},
  year={2026}
}

@inproceedings{zamir2020learning,
  title={Learning enriched features for real image restoration and enhancement},
  author={Zamir, Syed Waqas and Arora, Aditya and Khan, Salman and Hayat, Munawar and Khan, Fahad Shahbaz and Yang, Ming-Hsuan and Shao, Ling},
  booktitle={European conference on computer vision},
  pages={492--511},
  year={2020},
  organization={Springer}
}

@inproceedings{zamir2021multi,
  title={Multi-stage progressive image restoration},
  author={Zamir, Syed Waqas and Arora, Aditya and Khan, Salman and Hayat, Munawar and Khan, Fahad Shahbaz and Yang, Ming-Hsuan and Shao, Ling},
  booktitle={Proceedings of the IEEE/CVF conference on computer vision and pattern recognition},
  pages={14821--14831},
  year={2021}
}

@inproceedings{liang2021swinir,
  title={Swinir: Image restoration using swin transformer},
  author={Liang, Jingyun and Cao, Jiezhang and Sun, Guolei and Zhang, Kai and Van Gool, Luc and Timofte, Radu},
  booktitle={Proceedings of the IEEE/CVF international conference on computer vision},
  pages={1833--1844},
  year={2021}
}

@inproceedings{chen2021hinet,
  title={Hinet: Half instance normalization network for image restoration},
  author={Chen, Liangyu and Lu, Xin and Zhang, Jie and Chu, Xiaojie and Chen, Chengpeng},
  booktitle={Proceedings of the IEEE/CVF conference on computer vision and pattern recognition},
  pages={182--192},
  year={2021}
}

@inproceedings{wang2022uformer,
  title={Uformer: A general u-shaped transformer for image restoration},
  author={Wang, Zhendong and Cun, Xiaodong and Bao, Jianmin and Zhou, Wengang and Liu, Jianzhuang and Li, Houqiang},
  booktitle={Proceedings of the IEEE/CVF conference on computer vision and pattern recognition},
  pages={17683--17693},
  year={2022}
}

@inproceedings{zamir2022restormer,
  title={Restormer: Efficient transformer for high-resolution image restoration},
  author={Zamir, Syed Waqas and Arora, Aditya and Khan, Salman and Hayat, Munawar and Khan, Fahad Shahbaz and Yang, Ming-Hsuan},
  booktitle={Proceedings of the IEEE/CVF conference on computer vision and pattern recognition},
  pages={5728--5739},
  year={2022}
}

@inproceedings{chen2022simple,
  title={Simple baselines for image restoration},
  author={Chen, Liangyu and Chu, Xiaojie and Zhang, Xiangyu and Sun, Jian},
  booktitle={European conference on computer vision},
  pages={17--33},
  year={2022},
  organization={Springer}
}

@inproceedings{chu2022improving,
  title={Improving image restoration by revisiting global information aggregation},
  author={Chu, Xiaojie and Chen, Liangyu and Chen, Chengpeng and Lu, Xin},
  booktitle={European Conference on Computer Vision},
  pages={53--71},
  year={2022},
  organization={Springer}
}

@inproceedings{tu2022maxim,
  title={Maxim: Multi-axis mlp for image processing},
  author={Tu, Zhengzhong and Talebi, Hossein and Zhang, Han and Yang, Feng and Milanfar, Peyman and Bovik, Alan and Li, Yinxiao},
  booktitle={Proceedings of the IEEE/CVF conference on computer vision and pattern recognition},
  pages={5769--5780},
  year={2022}
}

@inproceedings{li2022all,
  title={All-in-one image restoration for unknown corruption},
  author={Li, Boyun and Liu, Xiao and Hu, Peng and Wu, Zhongqin and Lv, Jiancheng and Peng, Xi},
  booktitle={Proceedings of the IEEE/CVF conference on computer vision and pattern recognition},
  pages={17452--17462},
  year={2022}
}

@inproceedings{guo2024onerestore,
  title={Onerestore: A universal restoration framework for composite degradation},
  author={Guo, Yu and Gao, Yuan and Lu, Yuxu and Zhu, Huilin and Liu, Ryan Wen and He, Shengfeng},
  booktitle={European conference on computer vision},
  pages={255--272},
  year={2024},
  organization={Springer}
}

@inproceedings{radford2021learning,
  title={Learning transferable visual models from natural language supervision},
  author={Radford, Alec and Kim, Jong Wook and Hallacy, Chris and Ramesh, Aditya and Goh, Gabriel and Agarwal, Sandhini and Sastry, Girish and Askell, Amanda and Mishkin, Pamela and Clark, Jack and others},
  booktitle={International conference on machine learning},
  pages={8748--8763},
  year={2021},
  organization={PmLR}
}

@inproceedings{jia2021scaling,
  title={Scaling up visual and vision-language representation learning with noisy text supervision},
  author={Jia, Chao and Yang, Yinfei and Xia, Ye and Chen, Yi-Ting and Parekh, Zarana and Pham, Hieu and Le, Quoc and Sung, Yun-Hsuan and Li, Zhen and Duerig, Tom},
  booktitle={International conference on machine learning},
  pages={4904--4916},
  year={2021},
  organization={PMLR}
}

@inproceedings{chen2020simple,
  title={A simple framework for contrastive learning of visual representations},
  author={Chen, Ting and Kornblith, Simon and Norouzi, Mohammad and Hinton, Geoffrey},
  booktitle={International conference on machine learning},
  pages={1597--1607},
  year={2020},
  organization={PmLR}
}

@inproceedings{he2020momentum,
  title={Momentum contrast for unsupervised visual representation learning},
  author={He, Kaiming and Fan, Haoqi and Wu, Yuxin and Xie, Saining and Girshick, Ross},
  booktitle={Proceedings of the IEEE/CVF conference on computer vision and pattern recognition},
  pages={9729--9738},
  year={2020}
}

@inproceedings{caron2021emerging,
  title={Emerging properties in self-supervised vision transformers},
  author={Caron, Mathilde and Touvron, Hugo and Misra, Ishan and J{\'e}gou, Herv{\'e} and Mairal, Julien and Bojanowski, Piotr and Joulin, Armand},
  booktitle={Proceedings of the IEEE/CVF international conference on computer vision},
  pages={9650--9660},
  year={2021}
}

@article{dosovitskiy2020image,
  title={An image is worth 16x16 words: Transformers for image recognition at scale},
  author={Dosovitskiy, Alexey and Beyer, Lucas and Kolesnikov, Alexander and Weissenborn, Dirk and Zhai, Xiaohua and Unterthiner, Thomas and Dehghani, Mostafa and Minderer, Matthias and Heigold, Georg and Gelly, Sylvain and others},
  journal={arXiv preprint arXiv:2010.11929},
  year={2020}
}

@article{huang2017snapshot,
  title={Snapshot ensembles: Train 1, get m for free},
  author={Huang, Gao and Li, Yixuan and Pleiss, Geoff and Liu, Zhuang and Hopcroft, John E and Weinberger, Kilian Q},
  journal={arXiv preprint arXiv:1704.00109},
  year={2017}
}

@article{liu2026ntire,
 	title={NTIRE 2026 3D Restoration and Reconstruction in Real-world Adverse Conditions: RealX3D Challenge Results},
 	author={Liu, Shuhong and Bao, Chenyu and Cui, Ziteng and Chu, Xuangeng and Ren, Bin and Gu, Lin and Chen, Xiang and Li, Mingrui and Ma, Long and Conde, Marcos V and others},
 	journal={arXiv preprint arXiv:2604.04135},
 	year={2026}
}

@article{chang2026training,
 	title={Training-Free Model Ensemble for Single-Image Super-Resolution via Strong-Branch Compensation},
 	author={Chang, Gengjia and Ge, Xining and Yuan, Weijun and Li, Zhan and Song, Qiurong and Zhu, Luen and Liu, Shuhong},
 	journal={arXiv preprint arXiv:2604.11564},
 	year={2026}
}

@article{ge2026dual,
title={Dual-Branch Remote Sensing Infrared Image Super-Resolution},
 	author={Ge, Xining and Chang, Gengjia and Yuan, Weijun and Li, Zhan and Chen, Zhanglu and Yao, Boyang and Chen, Yihang and Deng, Yifan and Liu, Shuhong},
 	journal={arXiv preprint arXiv:2604.10112},
 	year={2026}
}

@article{chang2026beyond,
title={Beyond Model Design: Data-Centric Training and Self-Ensemble for Gaussian Color Image Denoising},
author={Chang, Gengjia and Ge, Xining and Yuan, Weijun and Li, Zhan and Song, Qiurong and Zhu, Luen and Liu, Shuhong},
journal={arXiv preprint arXiv:2604.11468},
year={2026}
}

@article{zheng20263d,
 	title={3D Smoke Scene Reconstruction Guided by Vision Priors from Multimodal Large Language Models},
 	author={Zheng, Xinye and Wang, Fei and Nie, Yiqi and Li, Kun and Chen, Junjie and Zhao, Jiaqi and Wei, Yanyan and Wu, Zhiliang},
 	journal={arXiv preprint arXiv:2604.05687},
 	year={2026}
}

@article{liu2026elog,
 	title={ELoG-GS: Dual-Branch Gaussian Splatting with Luminance-Guided Enhancement for Extreme Low-light 3D Reconstruction},
 	author={Liu, Yuhao and Wang, Dingju and Zheng, Ziyang},
 	journal={arXiv preprint arXiv:2604.12592},
 	year={2026}
}

@article{cao2026gensmoke,
 	title={GenSmoke-GS: A Multi-Stage Method for Novel View Synthesis from Smoke-Degraded Images Using a Generative Model},
 	author={Cao, Qida and Hu, Xinyuan and Shi, Changyue and Ding, Jiajun and Yu, Zhou and Yu, Jun},
 	journal={arXiv preprint arXiv:2604.03039},
 	year={2026}
}

@article{zhu2026naka,
 	title={Naka-GS: A Bionics-inspired Dual-Branch Naka Correction and Progressive Point Pruning for Low-Light 3DGS},
 	author={Zhu, Runyu and Dong, SiXun and Zhang, Zhiqiang and Ye, Qingxia and Xu, Zhihua},
 	journal={arXiv preprint arXiv:2604.11142},
 	year={2026}
}

@article{guo2026reliability,
  	title   = {Reliability-Aware Staged Low-Light Gaussian Splatting},
 	author  = {Guo, Haojie and Xian, Ke},
  	journal = {ResearchGate preprint},
 	year    = {2026}
}

@article{chen2026dehaze,
 	title={Dehaze-then-Splat: Generative Dehazing with Physics-Informed 3D Gaussian Splatting for Smoke-Free Novel View Synthesis},
 	author={Chen, Yuchao and Wang, Hanqing},
 	journal={arXiv preprint arXiv:2604.13589},
 	year={2026}
}

@article{fu2026smokegs,
  title={SmokeGS-R: Physics-Guided Pseudo-Clean 3DGS for Real-World Multi-View Smoke Restoration},
  author={Fu, Xueming and Han, Lixia},
  journal={arXiv preprint arXiv:2604.05301},
  year={2026}
}
}

\end{document}